\documentclass{article}

\usepackage{PRIMEarxiv}

\usepackage[utf8]{inputenc} 
\usepackage[T1]{fontenc}    
\usepackage[hyphens,spaces,obeyspaces]{url}
\usepackage{hyperref}
\usepackage{booktabs}       
\usepackage{nicefrac}       
\usepackage{microtype}      
\usepackage{lipsum}
\usepackage{fancyhdr}       
\usepackage{graphicx}       
\usepackage{tabularx}
\usepackage{multirow}

\usepackage{url}
\usepackage{xcolor}
\definecolor{newcolor}{rgb}{.8,.349,.1}
\usepackage{hyperref}
\usepackage{amsmath,amssymb,amsfonts}
\usepackage{algorithmic}
\usepackage[caption=false]{subfig}
\usepackage[export]{adjustbox}

\title{An Interpretable Deep-Learning Framework for Predicting Hospital Readmissions From Electronic Health Records}

\author{
  Fabio Azzalini \\
  Department of Electronics, Information\\
  and Bioengineering \\
  Politecnico di Milano \\
  Milan, Italy\\
  \texttt{fabio.azzalini@polimi.it}
\And
  Tommaso Dolci \\
  Lero -- the Science Foundation Ireland \\
  Research Centre for Software \\
  University of Limerick \\
  Limerick, Ireland\\
  \texttt{tommaso.dolci@lero.ie}
\And
  Marco Vagaggini \\
  Department of Electronics, Information\\
  and Bioengineering \\
  Politecnico di Milano \\
  Milan, Italy\\
  \texttt{marco.vagaggini@mail.polimi.it}
}

\begin{document}
\maketitle

\begin{abstract}
With the increasing availability of patient data, modern medicine is shifting towards prospective healthcare.
Electronic health records offer a variety of information useful for clinical patient characterization and the development of predictive models, given that similar medical histories often lead to analogous health progressions.
One application is the prediction of unplanned hospital readmissions, an essential task for reducing healthcare costs and improving patient outcomes.
While predictive models demonstrate strong performances especially with deep learning approaches, they are often criticized for their lack of interpretability, a critical requirement in the medical domain where incorrect predictions may have severe consequences for patient safety.
In this paper, we propose a novel and interpretable deep learning framework for predicting unplanned hospital readmissions, supported by NLP findings on word embeddings and by ConvLSTM neural networks for better handling temporal data.
We validate the framework on two predictive tasks for hospital readmission within 30 and 180 days, using real-world data.
Additionally, we introduce and evaluate a model-dependent technique designed to enhance result interpretability for medical professionals.
Our solution outperforms traditional machine learning models in prediction accuracy while simultaneously providing more interpretable results.
\end{abstract}


\section{Introduction}
With the increasing availability of extensive data on previously treated patients, modern medicine is progressively shifting towards \textit{prospective healthcare}, which aims to determine the risk for individuals for specific diseases, enable early detection and prevention, and maximize patient outcomes~\cite{snyderman2003prospective}.
In fact, the richness of healthcare data can be leveraged to predict health risks for incoming patients with high accuracy, bringing critical benefits to both healthcare structures and patients, by lowering hospitalization costs and improving the effectiveness of medical treatments.
\textbf{Electronic Health Records} (EHRs) are among the richest sources of health information.
This is evident in their steadily growing application, as reflected by the adoption of electronic health records EHRs by office-based physicians in the U.S. reaching 88\% in 2021~\cite{healthit}.
EHRs are composed of medical episodes, each containing a series of data about a patient, such as demographic information, biochemical profiles, past and present medical diagnoses, clinical notes, and more.
Each patient episode is associated with a timestamp, thus making the information contained in EHRs time-dependent.
Diseases often occur in interconnected patterns, forming clusters influenced by comorbidity occurring due to shared risk factors, biological mechanisms, or lifestyle influences~\cite{sharabiani2012systematic}.
This phenomenon enables EHRs to support a wide range of tasks in prospective healthcare, such as predicting heart failure~\cite{choi2016medical}, suicide risk~\cite{tran2015learning}, mortality~\cite{antikainen2023transformers}, or future diagnoses~\cite{[33]}.
However, information contained in EHRs is also very heterogeneous in nature: the presence of clinical codes (e.g., ICD-9 codes) and datetime objects (e.g., time of discharge) makes EHRs rather challenging to use effectively~\cite{shickel2017deep}.

One of the many applications of prospective healthcare is the prediction of unplanned hospital readmissions.
Unplanned readmissions have a significant impact on healthcare systems by consuming many resources, while simultaneously exposing patients to additional health risks.
To reduce and discourage unplanned hospital readmissions, governing bodies go so far as to implement regulatory measures.
For instance, the U.S. Congress created the Hospital Readmissions Reduction Program~\cite{[24]} to reduce funding for hospitals with excessive readmissions, which is often regarded as a symptom of incorrect treatment or premature discharge~\cite{[5]}.
With the ever-increasing volume of clinical data and the continuous improvements in the predictive capabilities of deep learning algorithms, it is now possible to develop more effective decision support systems for predicting hospital readmissions and further reducing patient risks.
However, a major concern with the introduction of deep learning models in healthcare is their lack of transparency and interpretability, leading to frequent criticism of their results~\cite{shickel2017deep,xiao2018opportunities}.
Particularly in the medical field, where incorrect predictions can severely affect patient health, it is crucial for predictive models to be sufficiently transparent and interpretable, allowing practitioners to understand the rationale behind decisions and to avoid both ethical and practical risks~\cite{khan2023drawbacks,watson2019clinical,criscuolo2022towards}.

Recently, solutions based on Convolutional Neural Networks (CNN) with few layers proved to be accurate in readmission prediction tasks, while their simple architecture also offers significant interpretability benefits~\cite{[1]}.
While more complex architectures, such as Transformers~\cite{vaswani2017attention}, achieve state-of-the-art results in many applications, their computational cost can make model training and inference challenging.\footnote{
Transformer architectures are computationally intensive due to the quadratic complexity of their self-attention mechanism, which scales as $O(n^2 \cdot d)$ with respect to the input sequence length $n$ and embedding dimension $d$. While efficient attention variants have been proposed to reduce this complexity~\cite{choromanskirethinking,dao2022flashattention,katharopoulos2020transformers,wang2020linformer}, their robustness is debated~\cite{keles2023computational}. Additionally, the large number of parameters in Transformer-based models contributes significantly to both memory usage and training/inference cost~\cite{hoffmann2022training,kaplan2020scaling}. Although they are highly parallelizable and there exist techniques to improve memory usage~\cite{kitaev2020reformer,korthikanti2023reducing}, these advantages may not fully compensate for their computational demands in resource-constrained domains such as healthcare.
}
Additionally, their interpretability relies on the often debated explainability of attention mechanisms~\cite{bibal2022attention,mohebbi2024transformer}.
Simpler models like CNNs reduces computational complexity while also providing transparent and interpretable results when combined with saliency methods~\cite{bastings2020elephant}, which is particularly important in healthcare applications.
However, a critical limitation of CNNs for predicting hospital readmission is the inability to properly assess long-term dependencies, which are particularly important for patients with a long clinical history~\cite{li2020behrt}.
To overcome this limitation, in this paper we introduce a novel framework for readmission prediction, named \textbf{ConvLSTM1d}.
ConvLSTM1d retains all the advantages of CNNs, such as low complexity and enhanced interpretability, while also incorporating a memory component through its ConvLSTM architecture, which leverages Long Short-Term Memory (LSTM) units to effectively capture long-term dependencies.
Our solution improves predictions of unplanned hospital readmissions compared to traditional machine learning models, while simultaneously boosting interpretability with respect to more complex deep learning architectures.
Until now, ConvLSTM models have mostly been used in the healthcare field for image segmentation tasks~\cite{azad2019bi,kang2022renal}.
Few notable works pioneered the use of ConvLSTM in conjunction with EHRs, but without focusing on the interpretability of the results~\cite{lin2018early,tian2024enhancing}.
In this paper, we also introduce and evaluate a model-dependent interpretability approach to visualize predictions, making them more understandable and accessible to medical practitioners.

This paper is organized as follows: in Section~\ref{sec:soa} we present works related to our solutions, regarding readmission prediction and interpretability in deep learning models; in Section~\ref{sec:model} we illustrate the architecture of ConvLSTM1d; Section~\ref{sec:interpretability} describes the model-dependent technique to make ConvLSTM1d more interpretable; Section~\ref{sec:results} presents the experimental setting and the performance results of our model; finally, in Section~\ref{sec:conclusions} we sum up this work and our contributions.

\section{Related Work}
\label{sec:soa}
Traditional machine learning techniques to analyze data contained in EHRs (e.g., random forest~\cite{tong2016comparison} and logistic regression~\cite{mortazavi2016analysis}) often require heavy pre-processing and feature engineering.
These steps are time-consuming and require extensive user domain knowledge.
In recent years, deep learning techniques have been increasingly adopted for the analysis of EHRs, with the reduced human contribution required~\cite{shickel2017deep}.
As pointed out by \cite{[33]}, research in disease predictions can be divided into two areas: \textit{specific-purpose progress modeling} relates to a specific type of disease, requiring heavy use of domain-specific knowledge; \textit{general-purpose progress modeling} is instead adaptable for a wide range of diseases.
We focus on research works from the second category, since our contribution is also general-purpose.

DeepCare~\cite{[2]} is an LSTM-based framework for outcome prediction from EHRs, used to model long-term dependencies in a variety of healthcare tasks.
Outcome prediction is a more general task compared to readmission prediction, but very similar in nature.
The authors evaluate DeepCare on three tasks: disease progression, intervention recommendation and readmission prediction (within 3 and 12 months).
Doctor AI~\cite{[33]} is another framework for outcome prediction, based instead on a Recurrent Neural Network (RNN).
Predictions of future diagnoses and medications are made by first taking, from the EHR, temporally-ordered codes for past diagnoses and medications, and then converting them into word embeddings to add a semantic dimension.
Doctor AI introduces a Top-K recall to mimic the behavior of doctors who would make a list of the K most likely future diagnoses and medications.
\cite{[5]} present a framework specifically designed to predict unplanned readmissions within 30 days from ICU discharge.
This system uses a large amount of patient data extracted from the MIMIC III database~\cite{[4]}.
The proposed model is based on LSTMs, to better handle time series data. In particular, the authors present a preliminary study to analyze and compare a bidirectional LSTM, a LSTM+CNN, and combinations of various features such as word embedding layers.
The best performances are obtained by an LSTM+CNN architecture using all the available data. 
At the time the paper was published, this type of architecture was very new, thus their results do not deviate much from a simple logistic regression, leaving room for improvement.
Finally, Deepr is yet another framework to solve outcome prediction tasks with EHRs~\cite{[1]}.
Not only it obtains accurate results, but it also represents a first attempt at creating an interpretable model that detects hidden relationships within the data, showing that interpretable deep learning has great potential in medical informatics.
Deepr is based solely on a CNN, unlike other frameworks which use RNNs and LSTMs. Despite a simple architecture, they find space for an embedding layer based on Word2Vec.
Moreover, the authors suggests that a convolutional layer is fundamental to determine the co-occurrence of diagnosis and procedures.

In the works previously presented, pre-trained embeddings are included to provide an additional layer for powerful semantic representations of medical concepts.
Models such as Word2Vec~\cite{mikolov2013distributed,mikolov2013efficient} can be trained on specific domains such as healthcare with little effort, to increase the accuracy of the word embeddings they produce.
In this context, the choice of the training data is crucial for obtaining good performance~\cite{[25]}.
For instance, \cite{[7]} introduce a pre-trained embedding model for representing medical concepts, based on Word2Vec and trained on a dataset containing information of over 4 million patients.
The generated vector space includes ICD-9-CM diagnoses and procedures codes, as well laboratory and drug codes from other medical dictionaries.
More recently, Transformer-based models such as BERT, which produce contextualized embeddings, have been included in risk prediction of patient hospital readmission~\cite{li2020behrt,arnaud2023explainable,rupp2023exbehrt}.
Unlike traditional methods that rely on static embeddings, which represent words in a fixed manner, BERT models generate embeddings that are dynamic and context-dependent, potentially offering a more flexible approach, at the expense of increasing complexity in terms of number of parameters, layers, and size of the training data.

Finally, the concept of interpretability is widely debated within the scientific community, and it largely depends on the task and context considered~\cite{[34]}.
In this paper, we consider as ``interpretable'' a model for which the reasons that led to a certain decision are made clear and accessible.
Recent literature reviews on the use of EHR for predicting hospital readmission emphasize the critical need to address interpretability in medicine~\cite{mahmoudi2020use}.
The evolution of this field necessitates the development and implementation of interpretable machine learning methods to enhance clinical usefulness.
LIME (Local Interpretable Model-agnostic Explanations)~\cite{[12]} is an example of a framework to make classiﬁers and regression models more interpretable and reliable.
LIME is very flexible and can be used for different types of input, such as images or text. 
Recent works from the literature, such as \cite{[26]}, studied the interpretability of CNNs applied to text, showing the limits of the traditional approaches that use convolutional filters as n-gram detectors, and a subsequent max-pooling layer to select the most relevant n-grams.
As for outcome prediction for EHRs with neural networks, the solution described by \cite{[14]} implements an easily interpretable model for predicting unplanned readmissions.
Similar to the Deepr framework, this classifier is built on a simple CNN architecture, which delivers relatively average performance but offers the significant advantage of being easily interpretable.

Recently, \cite{arnaud2023explainable} introduced a model of the BERT family for predicting patient admissions based on emergency department triage notes.
The authors leverage LIME to provide interpretable predictions through visualization of influential words in the notes, showing that combining NLP with explainable AI can assist in admission decisions.
BEHRT~\cite{li2020behrt} introduces a novel Transformer-based methodology for EHR data by combining embeddings—disease codes, age, positional info, and visit segments to pre-train the model on a masked language task.
BEHRT captures complex temporal and contextual dependencies, offering both interpretable results through attention visualization and higher predictive accuracy compared to traditional methods like RNNs and CNNs.
ExBEHRT~\cite{rupp2023exbehrt} extends the feature space of BEHRT to several multi-modal records, such as demographics, vital signs, and medications.
At the same time, interpretability of Transformer-based models is frequently debated in recent literature~\cite{mohebbi2024transformer}.
Whether attention mechanisms truly provide meaningful explanations in terms of faithfulness and plausibility is still an open question~\cite{bibal2022attention}.
Moreover, methods such as LIME can already boost interpretability of deep learning models, suggesting that attention visualization may not be needed~\cite{bastings2020elephant}.

Traditional machine learning methods remain a subject of study for hospital readmission prediction tasks, due to their ease of implementation, lower resource costs, and the favorable trade-off between transparency and performance.
\cite{davis2022effective} recently introduced two traditional models for readmission prediction, based on logistic regression and gradient boosting machine, that achieve good performance by integrating manually engineered features with Word2Vec features derived from administrative health data.

\section{ConvLSTM1d}
\label{sec:model}

The structural choices that motivated our research output, namely \textit{ConvLSTM1d}, are partly ascribable to the limitations highlighted by \cite{[1]} and their system, one of the best-performing readmission prediction frameworks present in the literature at the beginning of our research.
In their solution, long-term dependencies are captured through a max-pooling operation, which is a simplistic approach given the complex dynamics between care processes and disease processes.
A more effective model should prioritize time-sensitive information, where recent events are given more weight than distant ones.
Based on these statements, we came up with the following two solutions:
\textit{(i)} structure medical records in a new way, in order to take care of event timing, and \textit{(ii)} use a new model based on an LSTM architecture that is able to exploit this new structure to ``understand" long-term dependencies.

Our framework consists of two main steps: \textit{EHR representation} and \textit{readmission prediction}.

\subsection{EHR Representation}
The EHR representation is a crucial part of our methodology. 
The input of this phase is a data source in which each line uniquely corresponds to a hospital admission, and at the end of this phase the system returns a structured representation of the medical information to be used by the predictive algorithm.
To predict a future readmission, the whole patient history plays an important role. In this context, if an individual has multiple hospital admissions, to construct a useful training set, the patient's hospitalization at time $T$ must consider all the information about previous timestamps.
To include information on time intervals, the system computes the time gap between two consecutive admissions. We define five possible intervals: \textit{0-1 month, 1-3 months, 3-6 months, 6-12 months, and more than 12 months}.

Each hospitalization is described by a series of diagnoses and procedures, encoded within the dataset using ICD-9-CM codes, an international classification system for diseases, injuries, surgeries, and diagnostic and therapeutic procedures.\footnote{\url{https://www.cdc.gov/nchs/icd/icd9cm.htm}}
The following sequence of codes shows the example of a new admission of a patient with three prior admissions in her/his medical history; please note in italic the codes representing the time gaps between one admission and the next.

\begin{center}
\fbox{
\begin{minipage}{24em}
1910 Z83 911 1008 D12 K31 \textit{1-3m} R94 H53 Y83 M62 Y92 E87 T81 1893 D12 S14 738 1910 Z83 \textit{0-1m} T91 Y83 Y92 K91 M10 E86 \textit{6-12m} K31 1008 1910 Z13 Z83
\end{minipage}}  
\end{center}

The ICD-9-CM codes for each patient record are randomly shuffled within each admission. The reasons are twofold: \textit{(i)} given a single admission, there is no temporal order of the codes within the dataset; \textit{(ii)} as introduced by \cite{[1]}, shuffling records allows to capture many more potential clinical motifs and learn more combinations of codes.

Unlike a number of readmission prediction frameworks available in the literature, a crucial property of our system is the ability to understand long-term dependencies, which are of paramount importance to identify patterns in subsequent hospital admissions~\cite{li2020behrt}.
To achieve this goal, we introduce a novel representation of EHRs.
In our framework, data are organized in the following way:
each patient hospital admission takes into account, in a cumulative way, also the previous admissions.
The latter, however, are divided according to the time order in which they happened. Not only we consider the time gaps between two consecutive admissions, but, contrary to other methods, we also include this piece of information in a separate vector. 
\begin{figure}[t]
\centering
\subfloat[EHR first admission\label{fig:m7}]{
  \includegraphics[width=0.33\linewidth]{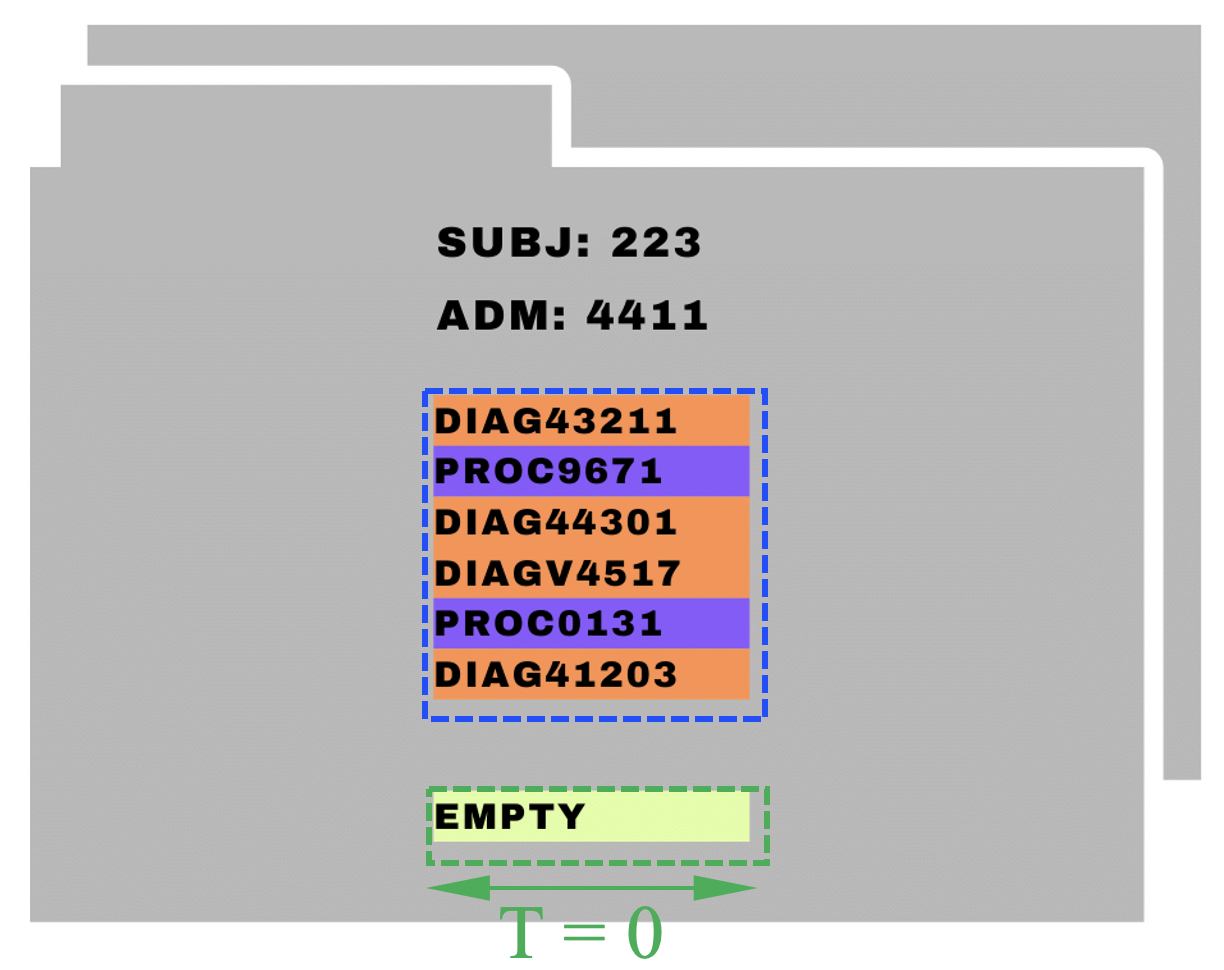}
}\hfil
\subfloat[EHR second admission\label{fig:m8}]{
  \includegraphics[width=0.33\linewidth]{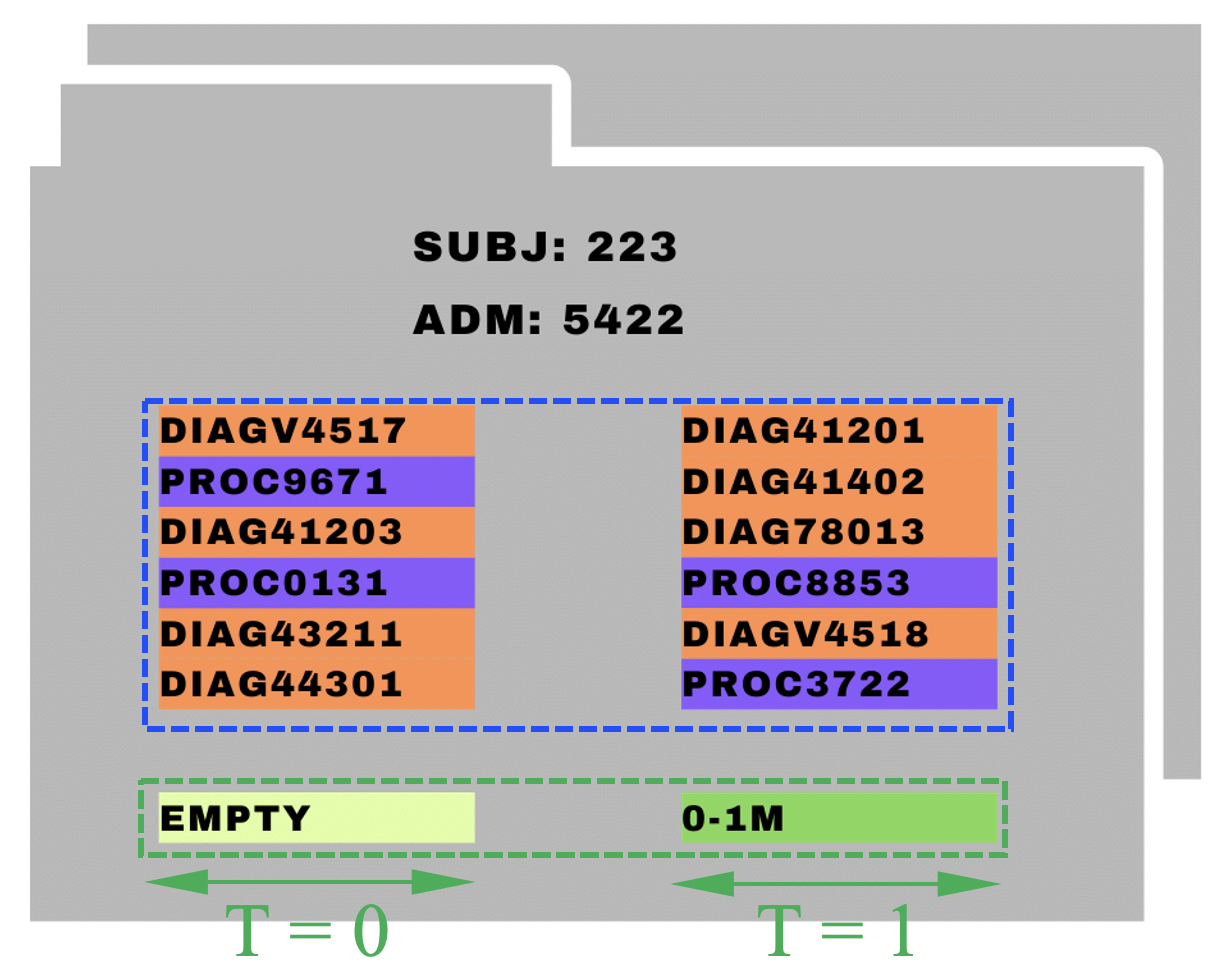}
}
\caption{Representation in ConvLSTM1d framework of the clinical history regarding the first and second admission of a patient.}
\label{fig:m_7_8}
\end{figure}
For greater clarity, in Fig.~\ref{fig:m_7_8} we display an example of how we represent admissions extracted from EHRs.
The vector representing the codes contained in all admissions undergone by the patient up to that point in time is surrounded by a blue dashed line, while the green dashed lines highlight the vectors corresponding to the time gaps between consecutive admissions.
The first value is always ``empty'' as it is relative to the patient's first admission.

Structuring the medical records in this way allows to take into account all the admissions related to the patient with the corresponding codes, without the necessity to make any split of the past data, typically by employing a sliding window.
This, along with the deep learning architecture adopted, allows to correctly assess the contribution of past events on future hospital readmissions.
Finally, the adopted EHR representation offer a high degree of flexibility, by enabling the integration of additional medical information into the prediction process. To incorporate new data, it is sufficient to encode it into a separate vector, ensuring the temporal order is preserved.
For instance, this could be the inclusion of drugs administered during each admission or patient-specific data such as age or weight from previous admissions.
Currently, since the other state-of-the-art methods do not offer this feature, we limit ourselves to using only procedural and diagnostic data with time gaps.

\subsection{Readmission Prediction Model}
\begin{figure}[tb]
    \centering
    \includegraphics[width=0.55\linewidth]{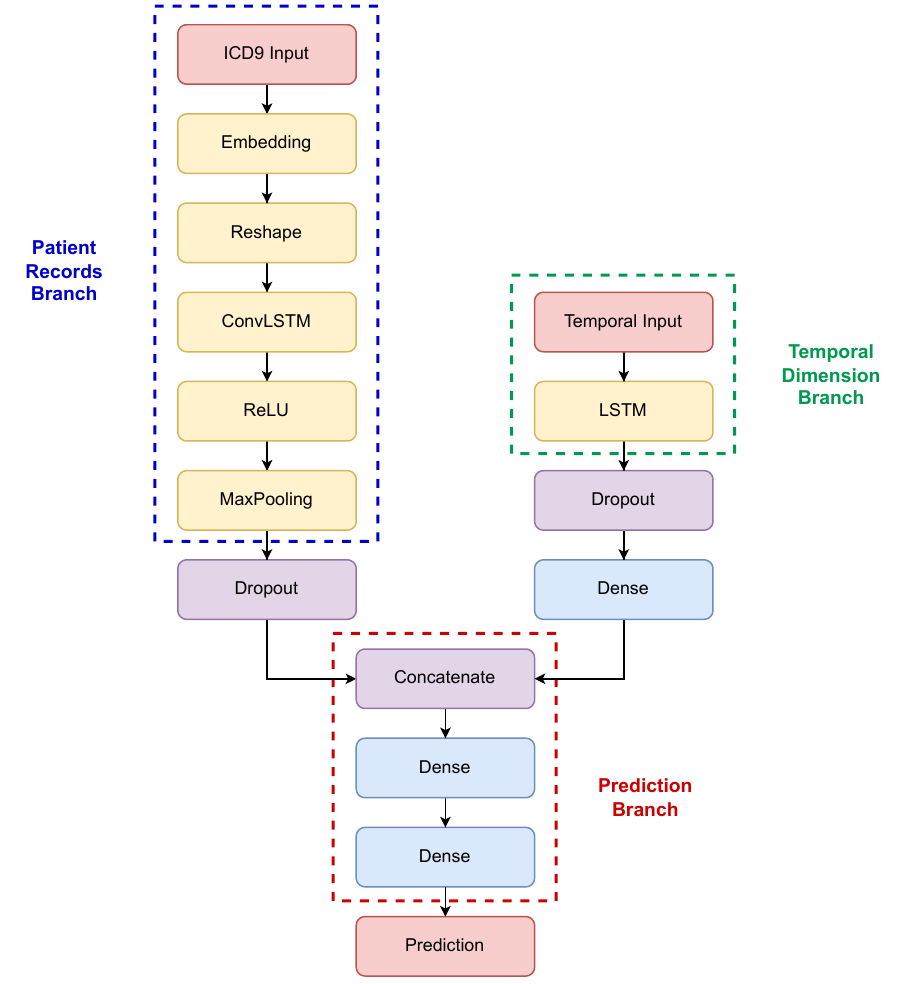}
    \caption{Layers implemented by ConvLSTM1d.}
    \label{fig:layers_conv}
\end{figure}

Our model is based on a neural network that receives as input two distinct vectors, the first containing the succession of patient hospital admissions (blue dashed box in Fig.~\ref{fig:layers_conv}), the other containing the corresponding time gaps at each admission (green dashed box in Fig.~\ref{fig:layers_conv}).
The core of our solution is represented by the Embedding layer and the ConvLSTM architecture.

\begin{figure}[t]
	\centering
	\includegraphics[width=0.65\textwidth]{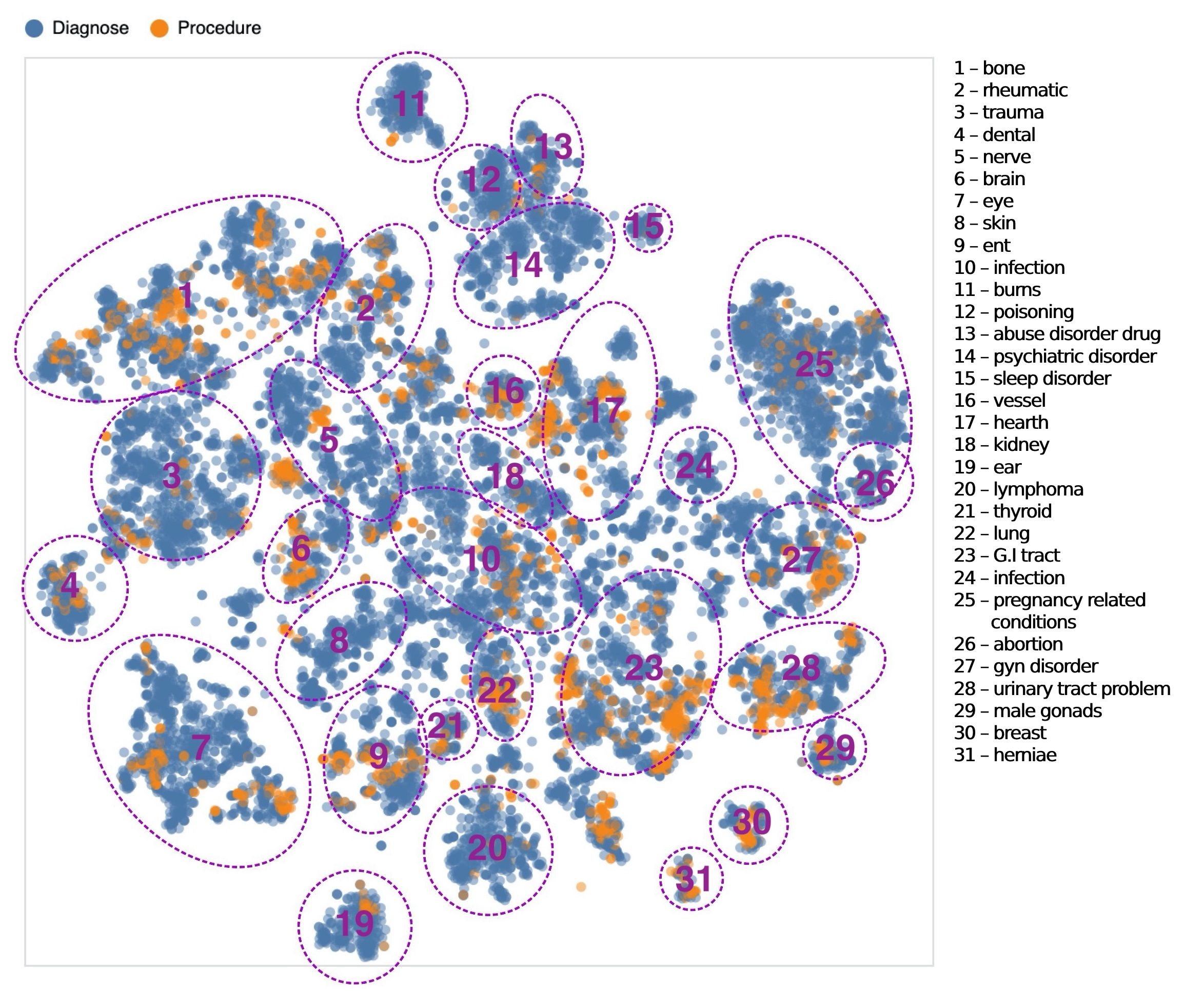}
	\caption{2d graphical representation of medical concepts by using t-SNE with ``cosine similarity'' as evaluation metric. Diagnoses are represented by blue points, procedures by orange points. Clusters are surrounded by a purple dashed line. On the right side the most prominent clusters are listed.
 }
	\label{fig:Embedding}
\end{figure}

The \textbf{Embedding layer} receives as input a patient medical record for a specific admission in the form of temporally divided ICD-9-CM codes, as represented in Fig.~\ref{fig:m_7_8}.
At this point, each code is converted into the corresponding unique word embedding vector, allowing us to better represent their semantics. 
This layer is necessary since computers work with numbers and cannot process strings in their native format.
To improve the performance of the system, we opted for a transfer learning approach. Specifically, the pre-trained word embedding was obtained by \cite{[7]}, who used the Word2Vec model with SkipGram implementation~\cite{mikolov2013efficient} to train on healthcare records of 4 million patients over a time span of 2-4 year.
The embedding vectors are 300-dimensional and contain many medical code standards, in our case we use only ICD-9-CM related to diagnoses and procedures.
By means of dimensionality reduction techniques such as t-SNE~\cite{van2008visualizing}, it is possible to represent the vectors generated by the word embedding model in 2 dimensions. 
This technique allows to graphically represent the semantically similar vectors as two neighboring points.
Fig.~\ref{fig:Embedding} shows the result produced by t-SNE, with cosine similarity as evaluation metric to estimate the semantic relatedness. 
In Fig.~\ref{fig:Embedding} we show ICD-9-CM codes of both diagnoses and procedures, colored in blue and orange respectively.
As evident from the graph, the interaction space of diagnoses and procedures is very complex, but it is possible to identify clusters defined by a specific medical domain.

After each code has been converted into the corresponding embedding vectors, the \textbf{ConvLSTM architecture} implements the forecasting algorithm.
This deep learning framework, introduced by \cite{[10]}, is an extension of LSTM, in which the matrix multiplication is replaced with the convolution operation.
As a result of its structure, this innovative architecture allows to exploit both the spatial and temporal information of the input, characteristics necessary for our system to consider both long-term dependencies between the admissions and abstract general health information (e.g., comorbidity) from sequences of specific procedures and diagnoses.
Following the ConvLSTM layer, max-pooling is performed, with the goal of making data depend only on the number of filters applied in the convolution and not on the initial input size. 
The second input contains a vector of categorical data (green dashed box in Fig.~\ref{fig:layers_conv}), equal in size to the time frames considered. The categories are those related to the various time gaps. Such data, being a simple sequence of mono-dimensional vectors, are processed by an LSTM layer, which permits to give the right weight to each event passed in the sequence. 
Subsequently, the outputs coming from the two branches of the network are concatenated (red dashed box in Fig.~\ref{fig:layers_conv}) and then a classifier with two Dense layers is used to predict whether a readmission will take place or not.


\section{Interpretability Approach}
\label{sec:interpretability}

Recent advancements in deep learning techniques have led to state-of-the-art results across various domains.
However, one of the most significant challenges associated with deep learning is the interpretability of their outcomes~\cite{watson2019clinical,khan2023drawbacks}.
These models are often regarded as black boxes, lacking internal transparency due to their complex network structure, which consists of multiple layers performing non-linear transformation.
In critical fields such as medicine, the interpretability of predictive tasks is of paramount importance, as a single decision can have profound implications on a patient's life.
A first attempt at improving the interpretability of deep learning models for readmission prediction was introduced by \cite{[1]}, based on the examination of convolution filters.
These filters act as n-gram detectors, which can be interpreted as ``clinical motifs''.
As discussed earlier in this paper, identifying clinical motifs is essential for accurately predicting unplanned readmissions.
However, as shown by \cite{[26]}, interpreting these filters can be challenging, as a single filter may capture multiple n-grams and potentially cancel the contributions of others.
In this context, a popular framework applicable to machine learning and deep learning models is LIME~\cite{[12]},
LIME approximates complex models with simpler, interpretable models in the local region of the prediction, allowing users to understand the factors influencing individual predictions.
In the literature, methods such as LIME are considered among the most viable way to improve interpretability in neural networks~\cite{bastings2020elephant}.

For this reason, we initially applied LIME to determine the contribution of each code in the EHR to the overall prediction.
This method works by perturbing the input data based on the concept of ``local fidelity'', offering a straightforward approach to identify the key factors influencing a specific prediction.
However, this method does not yield satisfactory interpretability results.
The challenge lies in the specific structure of the ConvLSTM1d model input, which consists of two temporally structured vectors—--one for ICD-9-CM codes and the other for time gaps---that do not allow for easy perturbation.
As a result, we developed a custom model-dependent interpretability approach designed to provide clearer insights when EHRs include temporal information.
The goal of this approach is to highlight the importance of each code in the final prediction, thereby assisting medical professionals in understanding the factors that most influence the outcome.
As highlighted by \cite{[14]}, this approach is feasible only because our prediction model's architecture consists of few layers, thus greatly improving its interpretability.
This would not be as effective with more complex architectures.
The goal of this approach is to obtain an easy-to-understand output structured as follows:

\begin{enumerate}
    \item Patient’s EHR, consisting of the various associated ICD-9-CM codes.
    \item The prediction results, in conjunction with the probability of each label.
    \item For each code in the medical record, the contribution of that code to the prediction.
\end{enumerate}

\noindent We now describe the approach step by step.
First, consider the output of a generic layer of the neural network: the correspondence between the output and the originating code is lost due to the max-pooling layer, which selects the largest response from each filter.
To obtain an interpretable result, we need to keep track, for each filter, of the input code that originates the largest response.
This allows to obtain a structure containing, for each input code, the correspondence between a filter and an ICD-9-CM code, named FILTER-CODE. 
It may happen that a code is not the maximum response of any filter, therefore not all the codes will be present in the FILTER-CODE structure.
The second step is to modify the intermediate results coming from the max-pooling layer.
The results of the max-pooling layer are necessarily $\geq 0$, since this layer is preceded by a ReLu activation function.
Consequently, by resetting the value of an intermediate output following the max-pooling layer, we ``deactivate'' the contribution of that specific filter in the final result. This property is particularly important to evaluate the importance of each filter on the final prediction. 

\begin{figure}[t]
	\centering
	\includegraphics[width=0.85\textwidth]{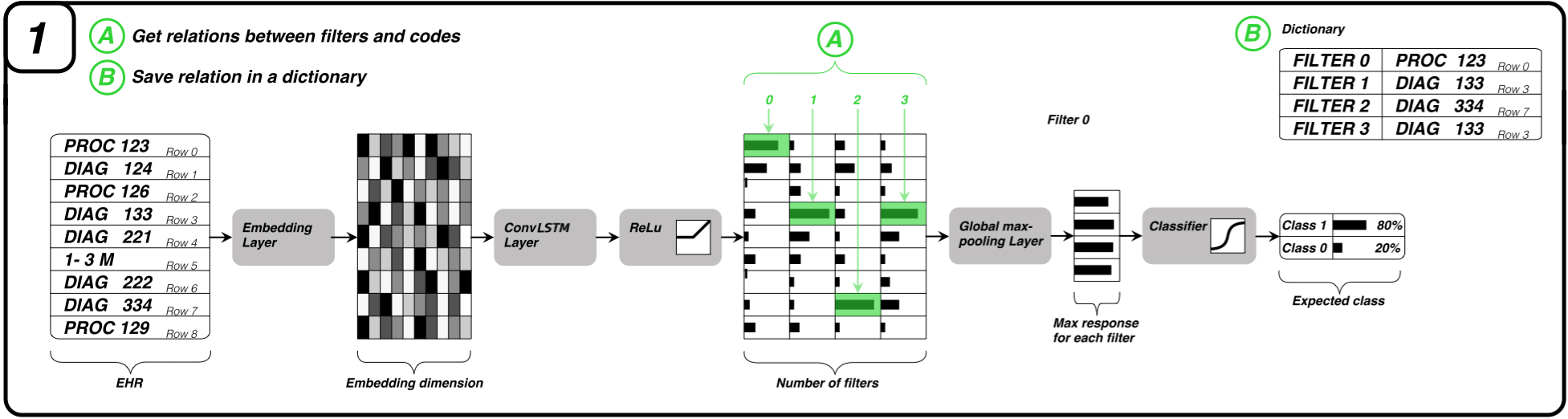}
	\caption{Implementation of the model-dependent interpretability approach -- first step.}
	\label{fig:int_diag1}
\end{figure}
\begin{figure}[t]
	\centering
	\includegraphics[width=0.85\textwidth]{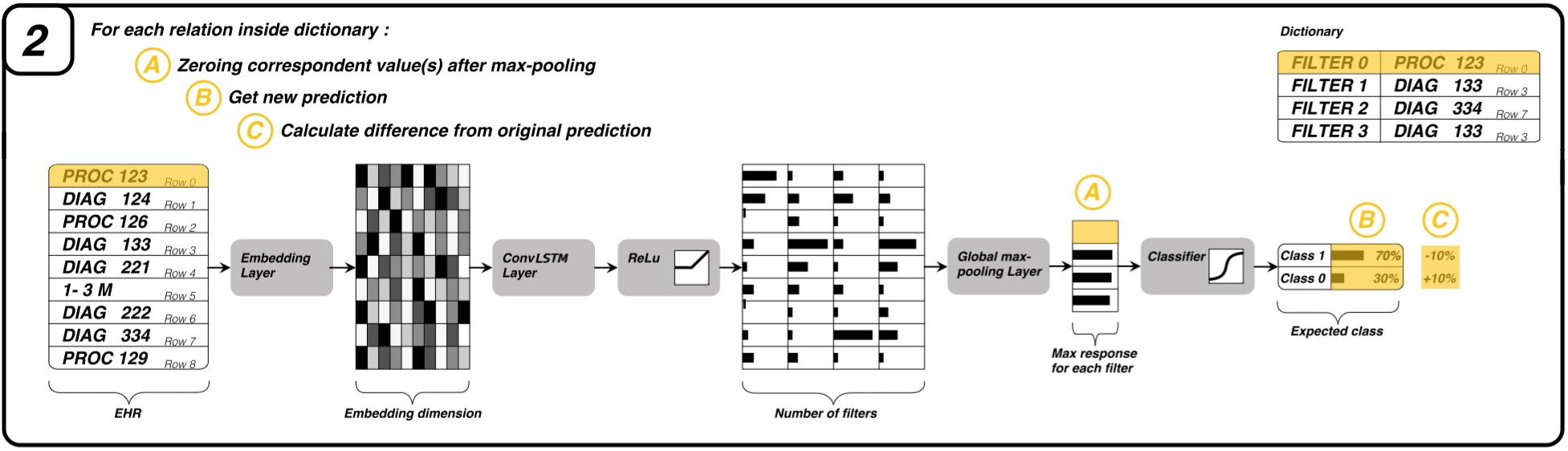}
	\caption{Implementation of the model-dependent interpretability approach -- second step.}
	\label{fig:int_diag2}
\end{figure}

For instance, to determine the contribution of code A123, we first search the structure containing FILTER-CODE matches to identify whether code A123 appears in one or more filters.
Suppose the following matches: A123-F1, A123-F2, A123-F3.
At this point, the intermediate results (i.e., the resutls after the max-pooling layer) corresponding to these three filters are set to zero.
By performing the prediction without these events in the EHR, it is possible to determine how much the predicted values change, thereby determining the contribution of code A123.
By repeating this procedure for each code, we obtain the interpretability of the whole EHR.

\begin{figure}[t]
	\centering
	\includegraphics[width=0.45\textwidth]{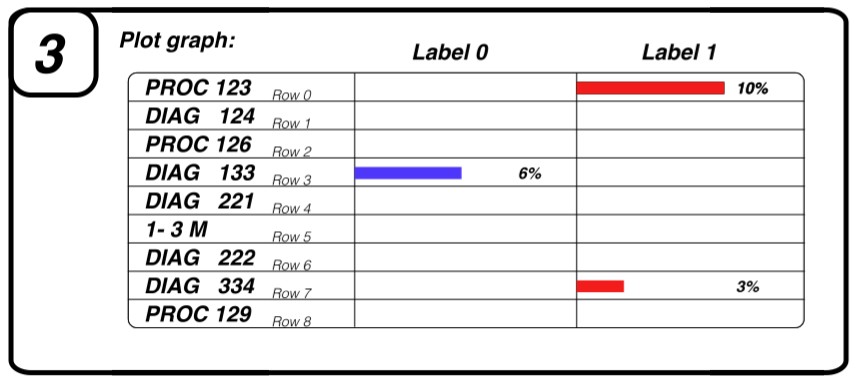}
	\caption{Implementation of the model-dependent interpretability approach -- third step.}
	\label{fig:int_diag3}
\end{figure}

To summarize, the contribution of a single code is evaluated with the following strategy: how much does the prediction change by zeroing out the internal activations related to the considered code?
Suppose the EHR has a prediction with Label 1 = 90\%.
Intuitively, if the prediction with Label 1 drops to 60\% when evaluating the contribution of the code A123, it means that A123 has a high contribution on this Label.
To quantify the contribution of a code we compute the difference of the two probabilities, i.e., $90-60 = 30\%$.
The reported contribution values must be considered as an indicative value that allows us to understand the reasons of such prediction, preventing a blind use of the model.
Fig.~\ref{fig:int_diag1},~\ref{fig:int_diag2} and~\ref{fig:int_diag3} visually describe the approach step by step.

\begin{figure}[t]
	\centering
	\includegraphics[width=0.48\textwidth]{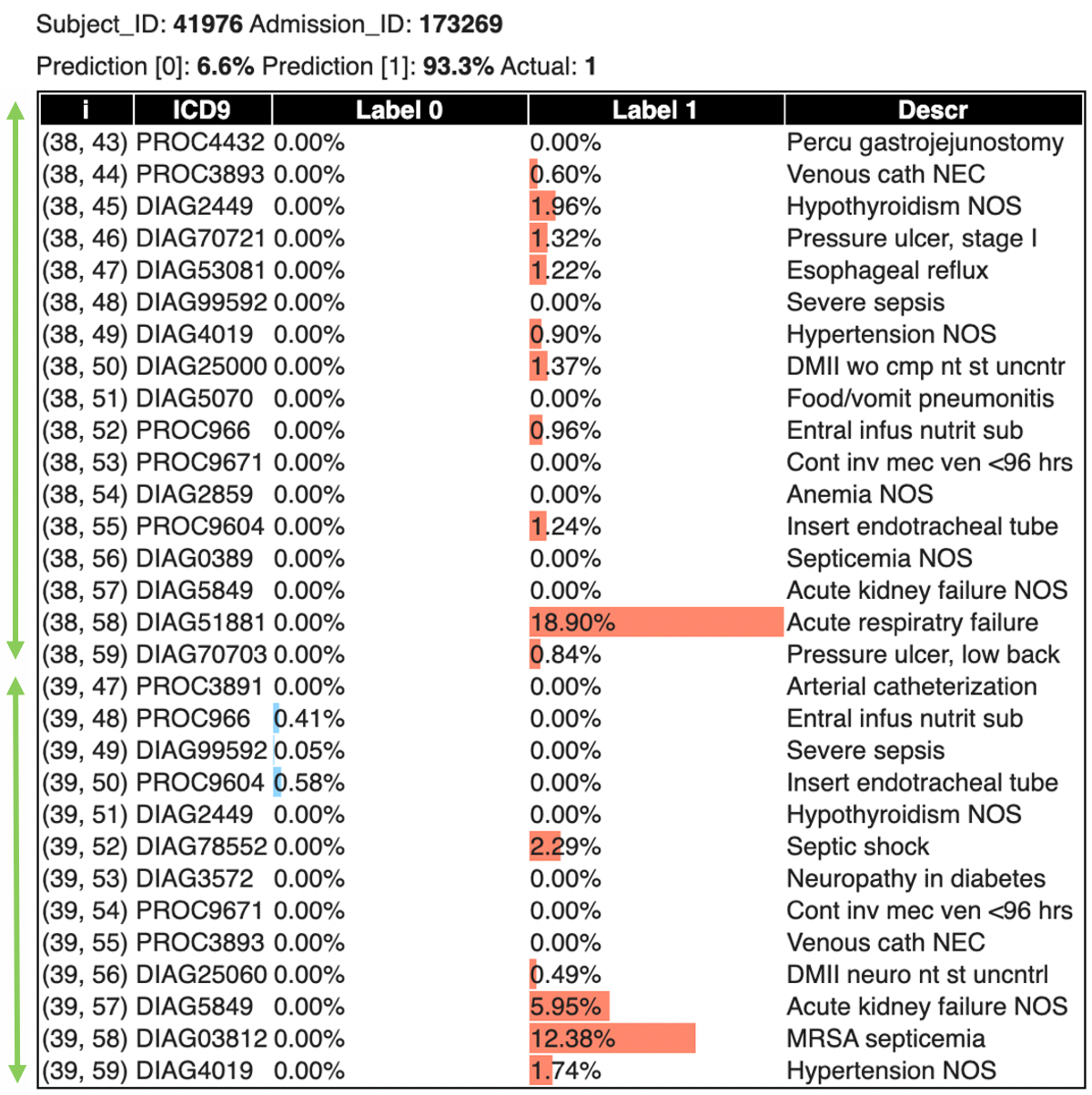}
	\caption{Interpretability of the model on a 30 days readmission prediction with model-dependent interpretability approach.}
	\label{fig:interpretability}
\end{figure}

Fig.~\ref{fig:interpretability} shows the result of the interpretability approach applied to a real-world example.
The two arrows represent two groups of temporally distinct codes (i.e., two consecutive admissions), thus allowing the model to leverage the previous information and ``weigh them temporally'' as a result of the memory of the ConvLSTM layer.
For each procedure and diagnosis, we report their importance on both the prediction of readmission (Label 1) and the prediction of no readmission (Label 0) within 30 days.
The results of this approach provide hints regarding which codes have the greatest influence on the prediction for a single admission.
By computing the average contributions across multiple admissions, it is possible to understand which procedures and diagnoses contribute most to the overall predictions.
Furthermore, through the use of charts and visualizations (e.g., Fig.~\ref{fig:int_30_180}), the model predictions can be easily explained to less technical individuals, such as clinicians.

\section{Experiments and Results}
\label{sec:results}
In this section, we first describe the training data used for ConvLSTM1d, then we report and discuss the performances on two prediction tasks: readmission within 30 and 180 days.
The main objective is to verify that our model improves performances in predicting unplanned hospital readmissions compared to state-of-the-art models, while at the same time allowing the interpretability of the results.
Moreover, we compare ConvLSTMd1 with Deepr~\cite{[1]} (a CNN-based model for readmission prediction) and with two traditional machine learning models, namely logistic regression and random forest.
The latter models are reported since they are widely used in hospital readmission prediction~\cite{huang2021application}.
For logistic regression and random forest, we use a bag-of-words model instead of word embeddings.
Bag of words is a simple yet popular approach in NLP problems, where text is represented as a set of words, without considering semantics or word order~\cite{jurafsky2000speech}.

\subsection{Training Dataset}
The EHRs used to train ConvLSTM1d are taken from the MIMIC III Database, or simply MIMIC (Medical Information Mart for Intensive Care)~\cite{[4]}.
MIMIC is a large, freely-available database comprising anonymized healthcare data associated with over 40,000 patients who stayed in intensive care units. 
The initial pre-processing of the dataset is fundamental to avoid errors in the data that may propagate to the subsequent steps. 
During pre-processing, we take inspiration from the procedure described by \cite{[5]}, that also use MIMIC III data for a task of readmission prediction.

First, we perform a data screening of MIMIC, removing all patients under the age of 18 and those who died in ICU, and correcting all inconsistencies.
We can see a patient's medical record made up of different hospital admissions (primary key: HADM\_ID), for each of which there may be different ICU admissions (primary key: ICUSTAY\_ID).
The tuples of the PROCEDURES and DIAGNOSES tables of the MIMIC III database are linked only to the single hospital admission, so we cannot know in which specific ICU a certain procedure has been performed.
Therefore, we only need the patients' procedures and diagnoses associated to a given hospital admission.
At this point, we have \textbf{35339} patients, with a total of \textbf{45357} hospital admissions.
Each tuple of the dataset corresponds to the hospital admission of one patient.
For the attribution of the label, we need to consider readmissions within 30 and 180 days.
The cases of positive labels are the following:
\begin{enumerate}
\item The patient was discharged but returned to the hospital within 1 month (\textbf{2689} admissions). This label is used for readmission prediction task within 30 days. 
\item The patient was discharged but returned to hospital within 6 months (\textbf{5764} admissions). This label is used for readmission prediction task within 180 days. 
\end{enumerate}

\begin{figure}[t]
     \centering
    \subfloat[30 days
         \label{fig:im8}]{
         \includegraphics[width=0.25\textwidth,valign=c]{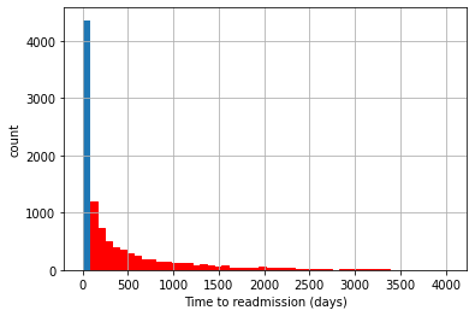}     
         }\hfil
     \subfloat[180 days
         \label{fig:im6}]{
         \includegraphics[width=0.25\textwidth,valign=c]{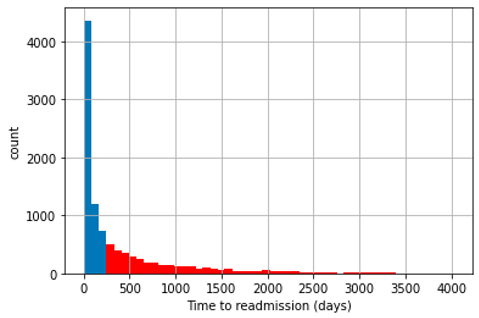}
         }\hfill
    \subfloat[30 days
         \label{fig:im11}]{
         \includegraphics[width=0.2\textwidth,valign=c]{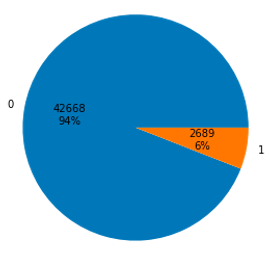}   
         }\hfil
    \subfloat[180 days
         \label{fig:im10}]{
         \includegraphics[width=0.2\textwidth,valign=c]{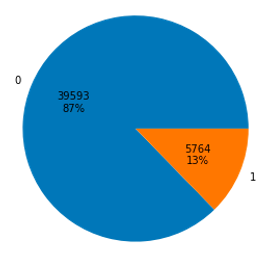}  
         }
    \caption{Data distribution in the training data. On the left, time to readmission in days: highlighted in red are those instances where number of days are more than 30 and 180 respectively. On the right, percentage of labels for readmissions within 30 and 180 days.}
    \label{fig:im_6_7_8_9_10_11}
\end{figure}

\noindent Fig.~\ref{fig:im8} and~\ref{fig:im6} report the data distributions from which we gathered this information.
As a result, the two datasets we are going to use have the following characteristics:
\begin{enumerate}
\item Readmission prediction within 30 days, totaling 5378 admissions, of which 2689 are positives (represented by the orange slice in Fig.~\ref{fig:im11}).
\item Readmission prediction within 180 days, totaling 11228 admissions, of which 5764 are positives (represented by the orange slice in Fig.~\ref{fig:im10}).
\end{enumerate}

\noindent Additionally, the two generated datasets are balanced. Otherwise, as evidenced by Fig.~\ref{fig:im11} and~\ref{fig:im10}, the negative label would be predominant in both tasks (with a prevalence of 94\% and 87\% respectively), and accuracy could not be used as a metric.
In both datasets, there may be multiple admissions for a single person.
The presence of multiple admissions is important to extrapolate as much information as possible from the rows of the datasets with a positive label (minority label).

\subsection{Training Parameters}

\begin{figure}[t]
    \centering
    \subfloat[loss
         \label{fig:p4}]{
         \includegraphics[width=0.33\textwidth]{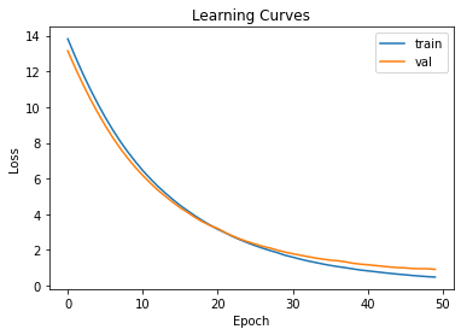}  
         }\hfil
    \subfloat[accuracy
         \label{fig:p2}]{
         \includegraphics[width=0.33\textwidth]{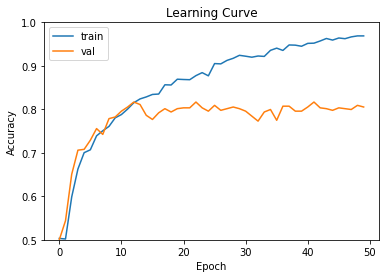}
         }
        \caption{Learning curves of loss and accuracy for  ConvLSTMd1 regarding the 30-days unplanned readmission task.}
        \label{fig:t_curves}
\end{figure}

After splitting the datasets into training, validation and test sets, we train our model using Adam optimizer and binary cross-entropy loss function.
Table~\ref{tab:param} reports the hyper-parameters used for the two tasks of 30-days and 180-days admission.
Fig.~\ref{fig:t_curves} represents the learning curves for both loss and accuracy on the 30-day unplanned readmission task.

\begin{table*}[t]
\centering
\small
\label{tab:param}
\begin{tabular}{ l | c | c }
\hline
\textbf{Parameter} & \textbf{30 days} & \textbf{180 days} \\ \hline
Number of filters  & 22 & 23 \\ \hline
Filter length   & 3 & 3 \\ \hline
Dropout rate  & 0.3 & 0.1 \\ \hline
Dense units  & 1000    & 700 \\ \hline
L2 convex regularization & 0.001 & 0.01 \\ \hline
L2 recurrent regularizer  & 0.1 & 0.001 \\ \hline
Recurrent dropout rate  & 0.2 & 0.2 \\ \hline
LSTM units & 20 & 30 \\ \hline
Recurrent dropout rate LSTM & 0.1 & 0.1 \\ \hline
Epochs & 19 & 30 \\ \hline
\end{tabular}
\caption{Hyperparameters used for ConvLSTM1d.}
\end{table*}

\subsection{Performance Metrics}
Before presenting the performance obtained by ConvLSTMd1, we carry out a brief summary of the main metrics adopted.
In the following description, TP stands for True Positive, TN for True Negative, FP for False Positive, FN for False Negative.
\textit{Accuracy} is an intuitive metric that consists in the ratio between the correctly predicted observations and the total number of observations. To obtain a good validity, the dataset must be balanced.
\textit{Accuracy} is an intuitive metric that measures the ratio of correctly predicted observations to the total number of observations. It is most informative when the dataset is balanced. Accuracy is useful to give an overall sense of performance.
\begin{displaymath}
Accuracy =\frac{T P+T N}{T P+T N+F P+F N}
\end{displaymath}

\noindent  \textit{Precision} is the ratio of correctly predicted positive observations to the total predicted positive observations. It is particularly useful when the cost of False Positives is high, since it focuses on the quality of positive predictions, minimizing the risk of incorrect positive classifications.
\begin{displaymath}Precision =\frac{T P}{T P+F P}
\end{displaymath}

\noindent \textit{Recall} is the ratio of correctly predicted positive observations to the total number of actual positive observations. It is particularly useful when the cost of False Negatives is high, since it favors capturing as many positive observations as possible, even at the risk of including False Positives.
\begin{displaymath}Recall =\frac{T P}{T P+F N}
\end{displaymath}

\noindent \textit{F1 Score} is the harmonic mean of precision and recall, providing a balanced measure that accounts for both False Positives and False Negatives. It allows the balance between precision and recall, offering a single metric that reflects both the quality and completeness of positive predictions.
\begin{displaymath}F_{1} = 2 \cdot \frac{\mathrm{p} \cdot \mathrm{r}}{\mathrm{p}+\mathrm{r}}
\end{displaymath}
\noindent where $p$ stands for Precision and $r$ for Recall.

\subsection{Results and Discussion}
Table~\ref{table:t_perf} illustrates the performance results of ConvLSTMd1 for both 30-days and 180-days unplanned readmission prediction.
Our model is trained to achieve the highest accuracy, which is used as the main evaluation metric.
To give a better overview, we compare the results of ConvLSTMd1 with two traditional machine learning approaches for data prediction (logistic regression and random forest) and with Deepr~\cite{[1]}, the most similar and advanced solution in literature for predicting hospital readmissions from EHRs.
All results are obtained from carrying out 10 separate training sessions for each task and considering the average score between sessions.
This expedient is necessary since deep learning models tend to obtain slightly different performance scores even for the same task.
\begin{table*}[tb]
\centering
\small
\label{table:t_perf}
\begin{tabular}{l cc }
\hline 
\textbf{Model} & \textbf{Readmission 30 days} & \textbf{Readmission 180 days} \\ \hline
\multirow{5}{*}{\textbf{Logistic Regression}} &
\begin{tabular}[t]{cccc} 
         Label&Precision&Recall&F1 \\  \hline
         0&0.739&\textbf{0.872}&0.800 \\ 
         1&\textbf{0.844}&0.691&0.760 \\ \\
         \multicolumn{3}{c}{Accuracy}  & 0. 782
\end{tabular} &
\begin{tabular}[t]{cccc} 
         Label&Precision&Recall&F1 \\ \hline
         0&0.757&\textbf{0.850}&0.801 \\ 
         1&0.829&0.728&0.775 \\ \\
         \multicolumn{3}{c}{Accuracy}  & 0.789 
\end{tabular}  
\\ \hline
\multirow{5}{*}{\textbf{Random Forest}}  &
\begin{tabular}[t]{cccc} 
         Label&Precision&Recall&F1 \\ \hline
         0&0.768&0.844&0.804 \\ 
         1&0.826&0.745&0.784 \\ \\
         \multicolumn{3}{c}{Accuracy} & 0.794 
\end{tabular}  & 
\begin{tabular}[t]{cccc}
         Label&Precision&Recall&F1 \\ \hline
         0&0.772&\textbf{0.850}&0.809 \\ 
         1&\textbf{0.833}&0.749&0.788 \\ \\
         \multicolumn{3}{c}{Accuracy} & 0.799 
\end{tabular}
\\ \hline
\multirow{5}{*}{\textbf{Deepr}} &
\begin{tabular}[t]{cccc c} 
         Label&Precision&Recall&F1 \\ \hline
         0&\textbf{0.802}&0.814&0.808 \\ 
         1&0.811&\textbf{0.799}&0.805 \\ \\
         \multicolumn{3}{c}{Accuracy} & 0.806 
\end{tabular}  & 
\begin{tabular}[t]{cccc} 
         Label&Precision&Recall&F1 \\ \hline
         0&0.783&0.841&0.811 \\ 
         1&0.828&0.767&0.796 \\ \\
         \multicolumn{3}{c}{Accuracy}  & 0.804 
\end{tabular} 
\\ \hline
\multirow{5}{*}{\textbf{ConvLSTM1d}} &
\begin{tabular}[t]{cccc} 
         Label&Precision&Recall&F1 \\ \hline
         0&0.801&0.848&\textbf{0.824} \\ 
         1&0.839&0.790&\textbf{0.813} \\ \\
         \multicolumn{3}{c}{Accuracy}  & \textbf{0.819} 
\end{tabular} &
\begin{tabular}[t]{cccc} 
         Label&Precision&Recall&F1 \\ \hline
         0&\textbf{0.818}&0.822&\textbf{0.820} \\ 
         1&0.821&\textbf{0.817}&\textbf{0.819} \\ \\
         \multicolumn{3}{c}{Accuracy}  & \textbf{0.820} 
\end{tabular} 
\\ \hline
\end{tabular}
\caption{Performance results from logistic regression, random forest, Deepr and ConvLSTM1d models, for both 30-days and 180-days unplanned readmission prediction. Highest results for each task in bold.}
\end{table*}
Considering the theory of classifiers performance, we are presented with two types of readmission errors:
\begin{enumerate}
\item The first case is when the patient is predicted to be ``not readmitted'' (label 0), but then it is readmitted, leading to a False Negative.
\item The second case is when the patient is predicted as ``readmitted'' (label 1), but later we discover that it is not readmitted, thus leading to a False Positive.
\end{enumerate}
From the patient's health point of view, one should certainly try to minimize False Negatives, trying to get a recall as high as possible. 
From the perspective of hospital costs in terms of both economics and available beds, we must consider that a False Positive is using resources that could be better spent.
Moreover, we need to consider that recall and precision represent a trade-off. In our case, since we are both interested in protecting the patient health, at the same time without wasting too many hospital resources that might be needed by other patients, the F1 score is the best metric to summarize the performance of our model.

From Table~\ref{table:t_perf} it is evident that ConvLSTM1d is the best model in terms of accuracy, substantially outperforming all the other models.
In particular, logistic regression and random forest obtain good results in either precision or recall for certain specific cases, but with overall accuracy and F1 scores significantly lower compared to our model.
Deepr considerably improves on both accuracy and F1 scores with respect to traditional machine learning methods, but it prevails over ConvLSTM1d only in recall for unplanned readmission within 30 days.
However, it should be noted that precision is much lower. Consequently, this is reflected in the F1 score, which favors ConvLSTM1d in both tasks.
On the other hand, ConvLSTM1d obtains higher results in both accuracy and F1 scores for both labels and both tasks.

\subsection{Model Interpretability Validation}
\begin{figure}[t]
\centering
  \includegraphics[width=0.38\linewidth]{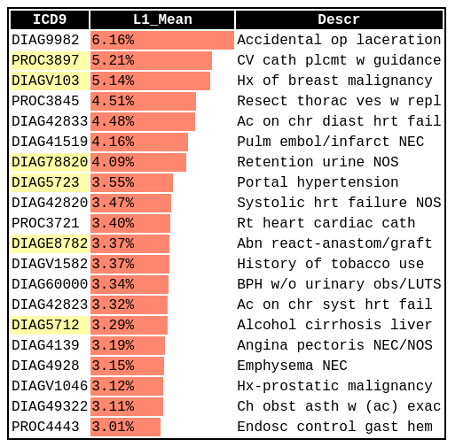}
  \hfil
  \includegraphics[width=0.38\linewidth]{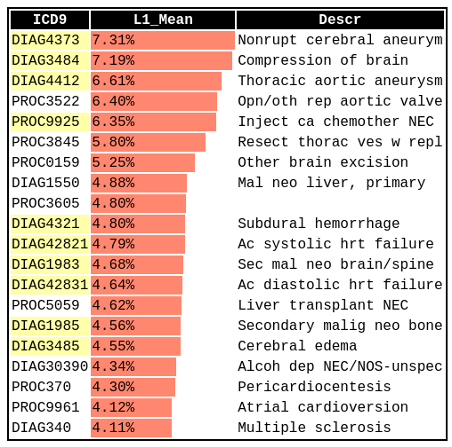}
\caption{ConvLSTM1d validation using interpretability results for 30 days (on the left) and 180 days (on the right) prediction tasks.}
\label{fig:int_30_180}
\end{figure}

In this section, we apply the interpretability approach introduced in Section~\ref{sec:interpretability}.
To validate the predictions made by our models, it is crucial to understand which procedures and diagnoses contribute most to the predictions on average.
To achieve this, we compute the average contributions across multiple admissions.
A similar approach can be used to evaluate the overall quality of the model by averaging the results across all admissions.
In this section, we present the results of the validation approach for both the readmission tasks considered in this paper, i.e., readmission within 30 and 180 days.
Fig.~\ref{fig:int_30_180} shows the results obtained. We report the average results for Label 1 (readmission) in both scenarios.
For the sake of brevity, we present the 20 most common codes in each category.

Analyzing the results of the 30-days readmission task, in first position is ranked the ICD-9-CM code related to the insertion of a central venous catheter related to several procedures in interventional and diagnostic/therapeutic settings (PROC 3897).
Liver disease, including portal hypertension (DIAG 5723) and alcoholic cirrhosis (DIAG 5712), also appears to have an important weight in short-term readmission.
Overall, the model has a high propensity for cardiovascular and respiratory diseases.
The most represented cancers are breast cancer (DIAG V103)---first type of cancer in terms of prevalence for women---and malignant prostate cancer (DIAG V1046)---first type of cancer for men. 
Considering the 180-days readmission task, among the most common manifestations are cardiac diseases (DIAG 42831, DIAG 42821), cerebral diseases (DIAG 3484, DIAG 4321), hepatic diseases (DIAG 5723, DIAG 5712), but also the injection of chemotherapeutic drugs for the treatment of neoplasms (PROC 9925).
Overall, codes related to the cardiac and the neurological domain are frequently present, and it can also be noted the importance attributed to multiple sclerosis (DIAG 340), which has a known fluctuating course with exacerbation. 
In summary, by analyzing the results of both 30 and 180 days readmission tasks, we can conclude that the cardiological field is important both in the short and long term, as well as the hepatic domain.
Invasive diagnostic and interventional procedures, as well as pulmonary problems, are very much present in the 30-day readmission scenario, while neurology is relevant only in 180-day readmissions. 

\section{Conclusions}
\label{sec:conclusions}
In this work, we addressed the problem of predicting unplanned hospital readmission from patients' EHRs.
We defined a novel neural network framework, named ConvLSTM1d, that outperforms existing traditional machine learning models while providing at the same time interpretable results.
Our model is based on a ConvLSTM architecture to implement a temporal dimension that allows to consider past events and the full disease history of the patient.
We tested ConvLSTM1d on two unplanned hospital readmission predictive tasks, within 30 and 180 days, using EHR data extracted from the large medical database MIMIC III.
Our solution achieves higher accuracy compared to both traditional machine learning (logistic regression and random forest) and Deepr, a CNN -based deep learning model for predicting hospital readmissions.
Additionally, our results represents, as far as we know, the first use of a ConvLSTM network with word embeddings for hospital readmission prediction, in conjunction with EHR data.
Finally, we presented and validated a visual method to improve the interpretability of our predictions, based on custom model-dependent approach.

In the future, we will consider the application of this study to a larger dataset. At the moment, MIMIC III is the only large public database available containing detailed information on patients' EHRs.
Moreover, time intervals and ICD-9-CM codes related to diagnoses and procedures represent a small portion of the information contained in EHRs: more information may be used to improve the predictive performance, such as the history of prescribed medications and laboratory tests for patients in intensive care units.
Currently, our solution is based on few layers. While this simplicity allows for a much more interpretable solution, we recognize that a more elaborate and deeper network may improve the performance of the model. However, this may come at the expense of reducing the overall interpretability of the system.
Therefore, experiments and comparisons with more complex architectures (e.g., Transformers) will be included in future work.
In particular, to achieve the double goal of improving performance while also keeping computational effort low, we plan to experiment with lightweight Transformers architectures such as Performer~\cite{choromanskirethinking} or Linformer~\cite{wang2020linformer}, and with a ConvLSTM network integrated with BERT-based language models for word vector encoding.
We plan to consider both fine-tuned medical models such as ClinicalBERT~\cite{huang2019clinicalbert} or BioBERT~\cite{lee2020biobert}, and distillation approaches for reducing the size of the model, like DistilBERT~\cite{sanh2019distilbert} and TinyBERT~\cite{jiao2020tinybert}.
Therefore, we plan to benchmark our current solution against attention-fueled solutions, both in terms of performance and interpretability, by evaluating the critical trade-off between complexity, accuracy and transparency in deep learning models.

\section*{Acknowledgments}
This work was supported, in part, by Science Foundation Ireland grant 13/RC/2094\_P2 and co-funded under the European Regional Development Fund through the Southern \& Eastern Regional Operational Programme to Lero -- the Science Foundation Ireland Research Centre for Software.

\bibliographystyle{unsrt}
\bibliography{biblio}  

\end{document}